\documentclass[11pt]{article}
\usepackage[preprint]{acl}
\usepackage{times}
\usepackage{latexsym}
\usepackage[T1]{fontenc}
\usepackage[utf8]{inputenc}
\usepackage{microtype}
\usepackage{inconsolata}
\usepackage{graphicx}
\usepackage{amsmath,amssymb}
\usepackage{booktabs,tabularx,array}
\usepackage{needspace}
\usepackage{placeins}
\usepackage{pifont}
\usepackage{xcolor}
\usepackage{colortbl}
\usepackage{fontawesome5}
\usepackage{tikz}
\usepackage{pgfplots}
\pgfplotsset{compat=1.18}
\usetikzlibrary{arrows.meta,positioning}
\newcolumntype{Y}{>{\raggedright\arraybackslash}X}

\definecolor{familyshade}{gray}{0.95}

\title{Can Large Language Models Extend Their Agentic Capabilities to Goal Pursuit in Open-Ended Multi-Turn Interaction?}

\author{
  \normalfont\large
  \textbf{Mingyu Chen}\textsuperscript{1,*}\quad
  \textbf{Jingkai Lin}\textsuperscript{1,*}\quad
  \textbf{Xiaofen Xing}\textsuperscript{1,\textdagger} \\
  \normalfont\large
  \textbf{Liankai Huang}\textsuperscript{1}\quad
  \textbf{Ruiqi Xiong}\textsuperscript{1}\quad
  \textbf{Xiangmin Xu}\textsuperscript{2,1,\textdagger} \\[0.45em]
  \normalfont\large
  \textsuperscript{1}School of EE., South China University of Technology, Guangzhou, China \\
  \normalfont\large
  \textsuperscript{2}Foshan University, Foshan, China \\
  \normalfont\large
  \{eemingyuchen, eelinjingkai\}@mail.scut.edu.cn \\
  \normalfont\large
  \{xfxing, xmxu\}@scut.edu.cn
}
\hypersetup{pdfauthor={Mingyu Chen, Jingkai Lin, Xiaofen Xing, Liankai Huang, Ruiqi Xiong, Xiangmin Xu}}

\begin{document}
\maketitle
\begingroup
\renewcommand{\thefootnote}{\fnsymbol{footnote}}
\footnotetext[1]{Equal contribution.}
\footnotetext[2]{Corresponding authors.}
\endgroup
\begin{abstract}
Goal pursuit in open-ended dialogue requires agents to adapt their actions to interaction feedback across successive exchanges. We investigate whether agentic capabilities developed for multi-step tool use carry over to dialogue organized through the model's native tool interface, and how this changes interaction behavior. We define \emph{Tool-Mediated Interaction} (TMI) as a training-free protocol that represents dialogue actions as tool calls and counterpart responses as tool results, maintaining a continuous reasoning--action--observation loop with retained reasoning and interaction history. We evaluate TMI across social interaction, emotional support, persuasion, and price negotiation using two models and 15,678 main-experiment episodes. Across both models, TMI improves social goal attainment, emotional-support preference, donation amounts, and buyer negotiation margins over conversational baselines with and without thinking. A Qwen ablation further shows that TMI outperforms ordinary conversation with retained reasoning on all four primary outcomes. Behavioral analyses associate the improvements with task-specific patterns in initial proposals and responses to interaction feedback. Alongside these gains, TMI reduces mean reasoning tokens per decision by 42.0--81.5\% relative to the thinking baseline and lowers mean cross-turn phrase reuse across all eight model--environment pairs. These findings show that interaction organization shapes both goal attainment and reasoning across dialogue turns.
\end{abstract}

\begin{figure}[t]
\centering
\includegraphics[width=\columnwidth]{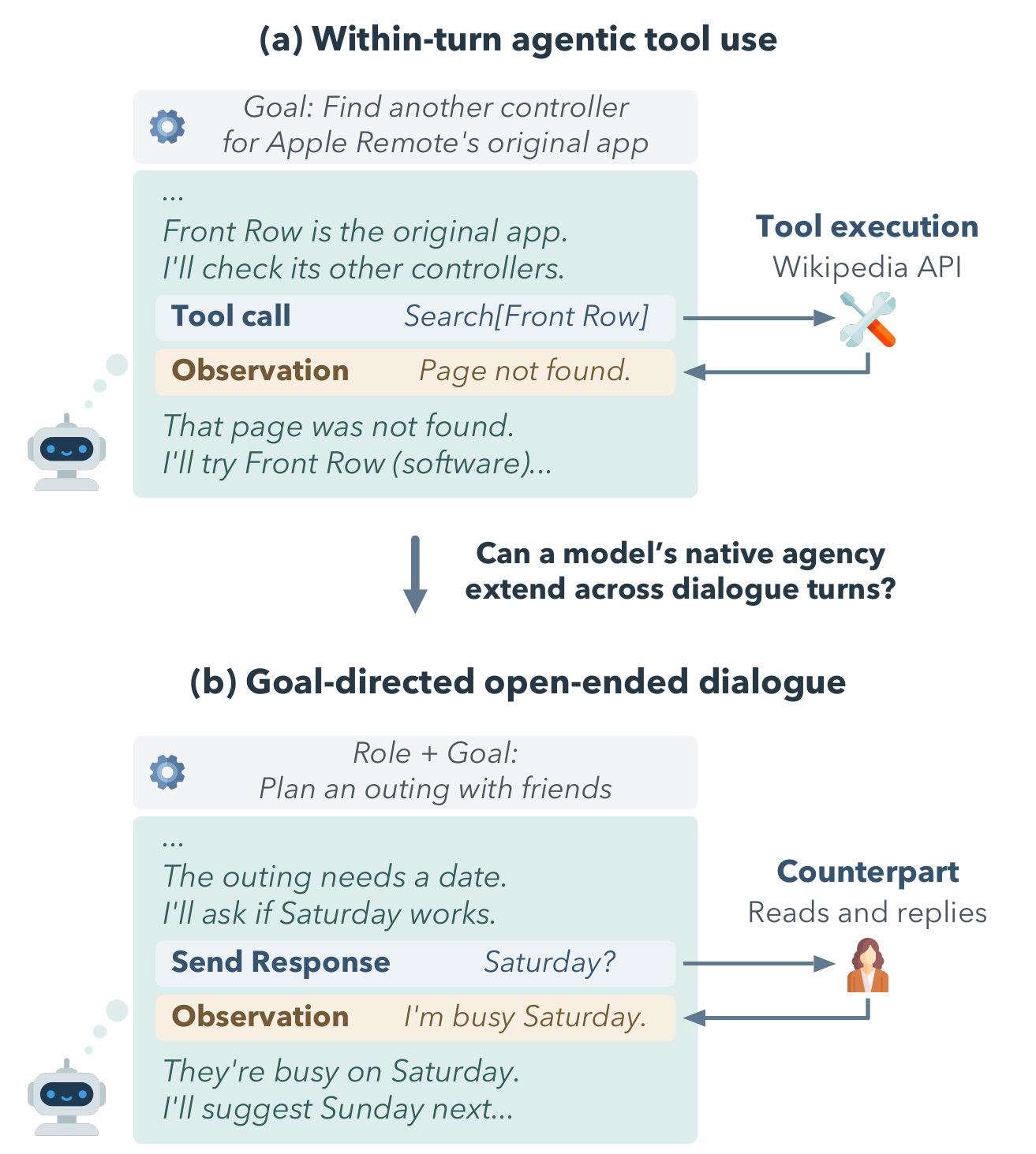}
\caption{Within-turn agentic tool use (a) and goal-directed multi-turn dialogue (b) share a reasoning--action--observation loop. Tool results and counterpart responses inform continued reasoning and subsequent actions toward a goal.}
\vspace{-0.8em}
\label{fig:motivation}
\end{figure}

\section{Introduction}
\label{sec:introduction}

Pursuing goals through dialogue requires adapting to the responses of other participants. In negotiation, persuasion, and interpersonal coordination, a proposal, refusal, or newly expressed preference can change which actions are feasible or useful. Intelligent agents must therefore remain guided by their goals and constraints while adjusting their actions as the interaction unfolds \citep{wooldridge1995intelligent}. This requires interpreting each response in relation to the overall objective and selecting subsequent actions accordingly.

To guide these decisions across turns, research on goal-directed dialogue has explored planning and strategic reasoning. Learned multi-turn planners steer generation toward dialogue-level objectives \citep{li2024dialogue}, while reinforcement learning improves dialogue policies through hindsight-regenerated interactions \citep{hong2024interactive} and utterance-level credit assignment informed by expected future outcomes \citep{feng-etal-2026-savoir}. Complementary approaches derive short- and long-term strategy guidance from expert dialogues \citep{yang-etal-2026-metro} or anticipate users' evolving emotional states to guide support strategies \citep{wu-etal-2026-emotion}. These approaches highlight the importance of connecting individual response choices to goal pursuit over successive exchanges.

In parallel, agentic post-training develops models' ability to pursue goals through multi-step tool use and environmental feedback \citep{lyu-etal-2026-agenticqwen}. Within a conversational turn, native tool execution allows a model to continue reasoning after each returned observation and select subsequent actions while retaining prior reasoning and observations \citep{xu2026deepseek}. This sustains a goal-directed reasoning--action--observation loop over successive tool calls. As illustrated in Figure~\ref{fig:motivation}, goal-directed dialogue has a similar structure: utterances serve as actions, and counterpart responses provide observations that inform continued reasoning and subsequent decisions. This correspondence motivates our central research question: \emph{can a model's native agency in tool use extend to goal-directed multi-turn dialogue, and how does this affect goal attainment and interaction behavior?}

To investigate this question, we study \emph{Tool-Mediated Interaction} (TMI), a training-free protocol that organizes dialogue through a model's native tool interface. In TMI, the model issues dialogue actions as tool calls and receives counterpart responses as tool results. This extends the native reasoning--action--observation loop across dialogue turns, with prior reasoning and interaction history retained to inform subsequent decisions. We compare TMI with dialogue conducted through assistant--user message exchanges to examine how interaction organization affects goal attainment and interaction behavior.

We evaluate TMI on \textsc{Sotopia} \citep{zhou2024sotopia}, ESC-Eval \citep{zhao-etal-2024-esc}, Persuasion for Good (P4G) \citep{wang-etal-2019-persuasion}, and CraigslistBargain (CB) \citep{he-etal-2018-decoupling}, using \textit{Qwen3.8-Flash} and \textit{DeepSeek-v4-Flash}. Across both models, TMI improves social goal attainment, emotional-support preference, donation amounts, and buyer negotiation margins relative to dialogue conducted through assistant--user messages, both with and without thinking. A reasoning-preservation control with Qwen further shows that TMI outperforms ordinary conversation with retained reasoning on all four primary outcomes. Trajectory analyses associate improved outcomes with actions that prepare for later progress and respond to feedback, including offers that leave bargaining room and inquiries that inform tailored support. Across all eight model--environment pairs, TMI also reduces mean reasoning tokens per decision by \textbf{42.0--81.5\%} relative to dialogue with thinking and lowers mean cross-turn phrase reuse.

Our contributions are threefold: (1) a controlled study of interaction organization in goal-directed dialogue, comparing TMI with assistant--user message exchanges; (2) evidence of improved primary task outcomes across four environments and two models, including a Qwen comparison with ordinary conversation that retains prior reasoning; and (3) analyses of interaction behavior and reasoning, identifying task-specific action patterns associated with improved outcomes and consistent reductions in reasoning length and cross-turn phrase reuse.

\begin{figure*}[t]
\centering
\includegraphics[width=\textwidth]{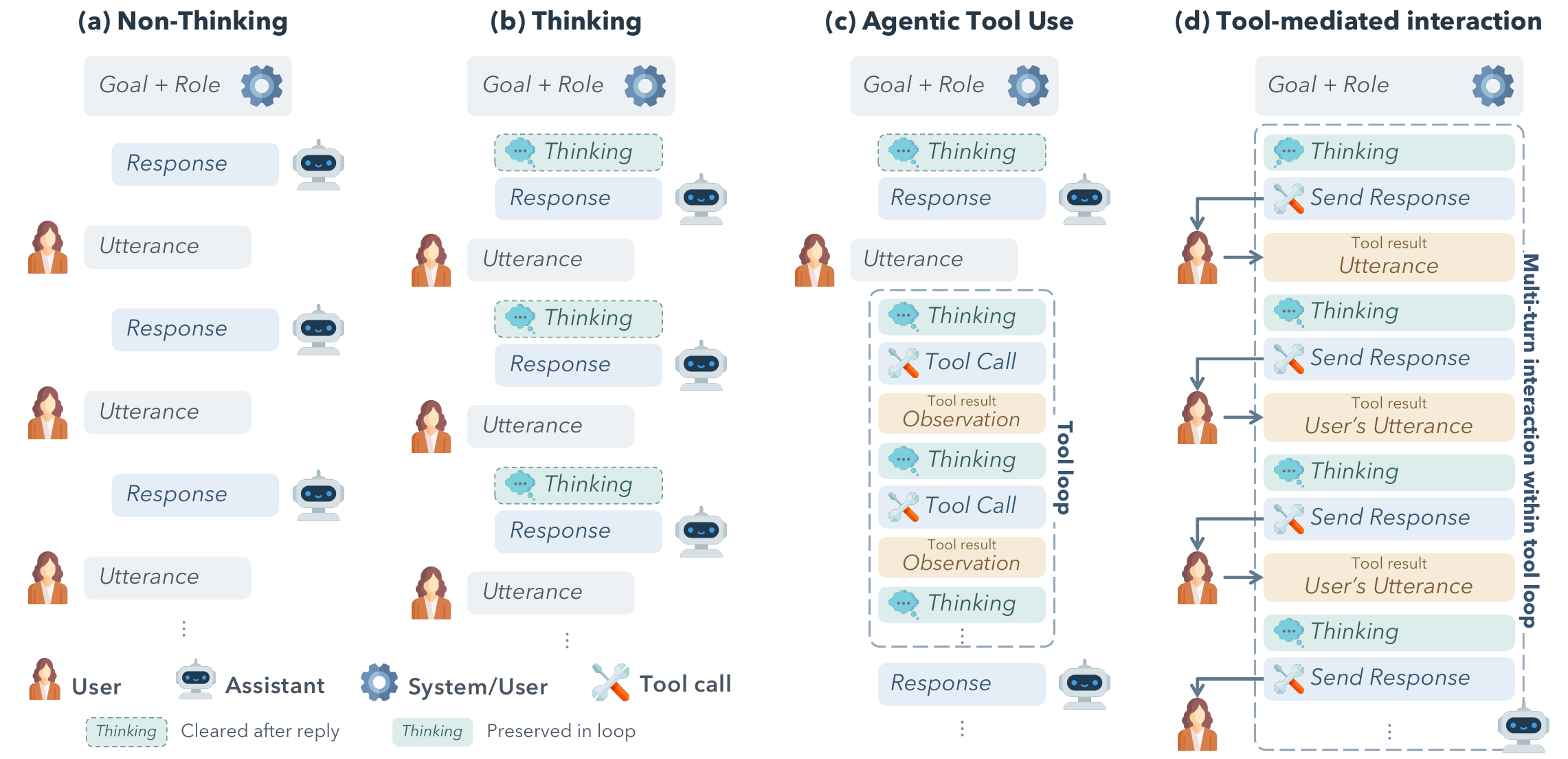}
\caption{From conversational responses to tool-mediated interaction. (a) The model responds directly; (b) it reasons before each response, with earlier reasoning omitted from subsequent requests. (c) Agentic tool use maintains a reasoning--action--observation loop during tool execution. (d) TMI applies this structure to dialogue: dialogue actions become tool calls and counterpart responses become tool results, preserving reasoning and interaction history across exchanges.}
\label{fig:framework}
\vspace{-0.8em}
\end{figure*}

\section{Related Work}
\label{sec:related}

\noindent\textbf{Goal-directed dialogue.} Goal-directed dialogue research has explored response planning, strategic guidance, and learning from multi-turn interaction. Prompted reasoning and lookahead search guide dialogue toward task goals \citep{deng-etal-2023-prompting,yu-etal-2023-prompt}, while learned planners and strategic reasoners provide trainable control over utterance generation \citep{deng2024plug,li2024dialogue,liu-etal-2025-epo}. Interaction-based learning uses dialogue outcomes to improve policies by filtering social trajectories, regenerating dialogues in hindsight, and attributing dialogue-level rewards to utterances \citep{wang-etal-2024-sotopia,hong2024interactive,feng-etal-2026-savoir}. Complementary approaches induce strategies and planning logic from expert dialogue transcripts \citep{yang-etal-2026-metro} or use emotional trajectories to anticipate future user states and guide support strategies \citep{wu-etal-2026-emotion}. Our study examines how interaction organization affects goal attainment, with model parameters, task goals, and permitted actions held fixed.

\noindent\textbf{Reasoning and interaction through tools.} Tool use connects language generation to external feedback through learned API invocation and interleaved reasoning--action trajectories \citep{schick2023toolformer,yao2023react}. Building on this connection, reinforcement learning develops tool-use policies for search, code execution, and broader agentic workflows, enabling models to learn when and how external operations contribute to task completion \citep{jin2025searchr,feng2026retool,lyu-etal-2026-agenticqwen}. Native tool-execution workflows can also retain reasoning and returned observations across successive tool calls, allowing subsequent decisions to build on accumulated execution history \citep{xu2026deepseek}. Tool use also extends to user interaction: $\tau$-bench evaluates dialogue combined with domain-specific APIs \citep{yao2024tau}, while PrefIx represents interaction behaviors as callable tools to evaluate preference adaptation, user experience, and task correctness \citep{li-etal-2026-prefix}. We study native tool execution for dialogue participants pursuing role-specific goals through interaction with a counterpart, examining their behavior and reasoning across exchanges.

\section{Tool-Mediated Interaction}
\label{sec:method}
\label{sec:formulation}

We study models pursuing assigned goals through open-ended, multi-turn dialogue. Figure~\ref{fig:framework} compares conventional dialogue with and without thinking to TMI, which organizes reasoning and interaction through native tool execution.

\subsection{Assistant--User Message Exchanges}
Let $g$ denote the participant's goal, $a_t$ its action at decision $t$, and $o_{t+1}$ the resulting observation. Actions include utterances and other permitted behaviors; observations include counterpart responses and visible environmental feedback. The model-visible context $C_t$ before decision $t$ comprises the initial context $C_1$ and retained interaction history. $C_1$ includes $g$, role information, interface instructions, and initial observations. Information access follows the environment's visibility rules.

Actions and observations are represented as assistant and user messages, respectively. The model generates $a_t$ directly in non-thinking mode (Figure~\ref{fig:framework}a), or a reasoning trace $r_t$ followed by $a_t$ in thinking mode (Figure~\ref{fig:framework}b):
\begin{equation}
\begin{aligned}
 a_t &\sim p_\theta(\cdot\mid C_t^{\mathrm{conv}}),\\
 (r_t,a_t) &\sim p_\theta(\cdot\mid C_t^{\mathrm{conv}}).
\end{aligned}
\label{eq:reasoning-action}
\end{equation}
Here $p_\theta$ denotes the model's generation distribution in the corresponding mode, with fixed parameters $\theta$. Both conditions retain visible dialogue history. The thinking condition omits prior reasoning traces from subsequent requests, yielding the shared context update
\begin{equation}
\begin{aligned}
 C_{t+1}^{\mathrm{conv}}=C_t^{\mathrm{conv}}
 &\oplus \mathsf{Assistant}(a_t)\\
 &\oplus \mathsf{User}(o_{t+1}),
\end{aligned}
\label{eq:conversation-update}
\end{equation}
where $\oplus$ concatenates messages in order, and $\mathsf{Assistant}$ and $\mathsf{User}$ construct messages with the corresponding roles.

\subsection{Dialogue through Native Tool Execution}
Agentic tool use supports a reasoning--action--observation loop before a conversational response (Figure~\ref{fig:framework}c), as exemplified by ReAct \citep{yao2023react}. TMI extends this loop across dialogue exchanges with a counterpart (Figure~\ref{fig:framework}d).

The model encodes each permitted action $a_t$ as an \texttt{interact} call, including its public utterance when applicable. The runtime executes the action, obtains the counterpart's response through a separate model request, and returns $o_{t+1}$ as the tool result. The counterpart responds according to its role and locally visible history; the model's reasoning remains private.

With thinking enabled, the model generates $(r_t,a_t)$ following the second line of Eq.~\ref{eq:reasoning-action}, conditioned on $C_t^{\mathrm{TMI}}$. Its context is updated as
\begin{equation}
\begin{aligned}
 C_{t+1}^{\mathrm{TMI}}=C_t^{\mathrm{TMI}}
 &\oplus \mathsf{Assistant}(r_t,\mathsf{Call}(a_t))\\
 &\oplus \mathsf{ToolResult}(o_{t+1}).
\end{aligned}
\label{eq:context-update}
\end{equation}
Here $\mathsf{Call}(a_t)$ denotes an \texttt{interact} invocation, and $\mathsf{ToolResult}$ constructs its result message. Reasoning, actions, and observations are retained across model requests until environment-defined termination. TMI changes the interface representation and retained reasoning history while keeping model parameters, goals, and permitted actions fixed.

\section{Experiments}
\label{sec:evaluation}

\subsection{Environments and Evaluation}
\label{sec:environments}
We evaluate TMI in four environments for goal-directed dialogue: social interaction, emotional support, persuasion, and price negotiation. We report benchmark scores and task-specific outcomes.

\noindent\textbf{Sotopia-Hard.} \textsc{Sotopia} simulates social interactions between participants with individual profiles and private social goals \citep{zhou2024sotopia}. We use \textsc{Sotopia-Hard}, a challenging subset covering negotiation, persuasion, and interpersonal coordination tasks \citep{wang-etal-2024-sotopia}, and evaluate both participant roles in separate episodes. We use Sotopia-Eval to assess goal achievement, believability, relationship, knowledge, secret keeping, social rule compliance, and financial and material benefits. Overall averages these seven dimensions.

\noindent\textbf{ESC-Eval.} ESC-Eval assesses emotional support in conversations between a supporter and a simulated seeker \citep{zhao-etal-2024-esc}. We evaluate the supporter using all 331 role cards in the English high-quality subset. ESC-RANK scores Fluency, Expression (diversity), Empathy, Information (suggestion effectiveness), Humanoid (humanlikeness), Skill (emotional support knowledge), and Overall (preference) on a 0--4 scale. Overall is scored independently.

\noindent\textbf{Persuasion for Good (P4G).} P4G concerns persuading a counterpart to donate part of a \$2 task payment to Save the Children \citep{wang-etal-2019-persuasion}. The evaluated agent acts as the persuader, interacting with a simulated counterpart assigned one of METRO's 200 evaluation personas \citep{yang-etal-2026-metro}. We report mean donation amount (\$0--\$2) and success rate, defined as the proportion of episodes ending in a positive donation.

\noindent\textbf{CraigslistBargain (CB).} CB concerns price negotiation over Craigslist listings \citep{he-etal-2018-decoupling}. We evaluate a buyer seeking a favorable price from a simulated seller, using METRO's 200 development assignments covering 85 listings \citep{yang-etal-2026-metro}. We report agreement rate and the buyer's normalized negotiation margin (utility). Let $t_b<t_s$ be the buyer and seller target prices and $m=(t_b+t_s)/2$. For an agreed price $p$, the buyer's margin is
\begin{equation}
\operatorname{Margin}_{b}=\frac{m-p}{m-t_b}.
\label{eq:cb-margin}
\end{equation}
Margins are not clipped, and episodes without agreement receive $0$. Success rate is the proportion of all episodes reaching agreement at $p\leq t_b$.

\begin{table*}[!t]
\begingroup
\centering\small
\setlength{\tabcolsep}{1.5pt}\renewcommand{\arraystretch}{1.18}
\begin{tabular*}{\textwidth}{@{\extracolsep{\fill}}lccccccccc@{}}
\toprule
& \multicolumn{5}{c}{Sotopia-Hard} & \multicolumn{4}{c}{ESC-Eval} \\
\cmidrule(lr){2-6}\cmidrule(lr){7-10}
Setting & Goal~$\uparrow$ & Rel.~$\uparrow$ & Know.~$\uparrow$ & Fin.~$\uparrow$ & Overall~$\uparrow$ & Expr.~$\uparrow$ & Info.~$\uparrow$ & Skill~$\uparrow$ & Overall~$\uparrow$ \\
\midrule
\rowcolor{familyshade}\multicolumn{10}{l}{\textit{Qwen3.8-Flash}} \\
w/o Thinking & 7.92\textsubscript{\textpm0.15} & 1.85\textsubscript{\textpm0.05} & 4.73\textsubscript{\textpm0.08} & 1.40\textsubscript{\textpm0.06} & 3.52\textsubscript{\textpm0.01} & 2.84\textsubscript{\textpm0.02} & 2.22\textsubscript{\textpm0.04} & 2.89\textsubscript{\textpm0.03} & 1.47\textsubscript{\textpm0.02} \\
w/ Thinking & \underline{8.19}\textsubscript{\textpm0.12} & \underline{2.14}\textsubscript{\textpm0.13} & \underline{5.15}\textsubscript{\textpm0.30} & \underline{1.49}\textsubscript{\textpm0.04} & \underline{3.70}\textsubscript{\textpm0.05} & \underline{2.93}\textsubscript{\textpm0.03} & \underline{2.34}\textsubscript{\textpm0.01} & \underline{2.95}\textsubscript{\textpm0.02} & \underline{1.69}\textsubscript{\textpm0.02} \\
TMI & \textbf{8.66}\textsubscript{\textpm0.04} & \textbf{2.71}\textsubscript{\textpm0.10} & \textbf{5.43}\textsubscript{\textpm0.32} & \textbf{1.65}\textsubscript{\textpm0.02} & \textbf{3.91}\textsubscript{\textpm0.06} & \textbf{2.96}\textsubscript{\textpm0.02} & \textbf{2.45}\textsubscript{\textpm0.02} & \textbf{2.98}\textsubscript{\textpm0.01} & \textbf{1.90}\textsubscript{\textpm0.003} \\
\midrule
\rowcolor{familyshade}\multicolumn{10}{l}{\textit{DeepSeek-v4-Flash}} \\
w/o Thinking & 8.02\textsubscript{\textpm0.10} & \textbf{2.48}\textsubscript{\textpm0.03} & \underline{5.33}\textsubscript{\textpm0.27} & \underline{1.55}\textsubscript{\textpm0.08} & \underline{3.73}\textsubscript{\textpm0.04} & 2.88\textsubscript{\textpm0.01} & 2.14\textsubscript{\textpm0.02} & 2.91\textsubscript{\textpm0.02} & 1.50\textsubscript{\textpm0.02} \\
w/ Thinking & \underline{8.09}\textsubscript{\textpm0.03} & 2.28\textsubscript{\textpm0.07} & 5.27\textsubscript{\textpm0.12} & 1.32\textsubscript{\textpm0.04} & 3.67\textsubscript{\textpm0.01} & \underline{2.94}\textsubscript{\textpm0.02} & \underline{2.22}\textsubscript{\textpm0.002} & \underline{2.96}\textsubscript{\textpm0.01} & \underline{1.72}\textsubscript{\textpm0.02} \\
TMI & \textbf{8.41}\textsubscript{\textpm0.13} & \underline{2.42}\textsubscript{\textpm0.06} & \textbf{5.34}\textsubscript{\textpm0.19} & \textbf{1.62}\textsubscript{\textpm0.03} & \textbf{3.80}\textsubscript{\textpm0.05} & \textbf{2.96}\textsubscript{\textpm0.01} & \textbf{2.25}\textsubscript{\textpm0.03} & \textbf{2.99}\textsubscript{\textpm0.003} & \textbf{1.92}\textsubscript{\textpm0.01} \\
\bottomrule
\end{tabular*}
\caption{Sotopia-Hard and ESC-Eval results (mean $\pm$ SD across three benchmark repeats). Rel.: Relationship; Know.: Knowledge; Fin.: Financial/Material Benefits; Expr.: Expression; Info.: Information. Bold and underline mark the best and second-best means within each model, ranked before rounding.}
\label{tab:social-support-main-results}
\endgroup
\par\vspace{6pt}
\begingroup
\centering\small
\setlength{\tabcolsep}{1.5pt}\renewcommand{\arraystretch}{1.18}
\begin{tabular*}{\textwidth}{@{\extracolsep{\fill}}lccccc@{}}
\toprule
& \multicolumn{2}{c}{P4G} & \multicolumn{3}{c}{CB Buyer} \\
\cmidrule(lr){2-3}\cmidrule(lr){4-6}
Setting & Donation (\$)~$\uparrow$ & Success (\%)~$\uparrow$ & Margin~$\uparrow$ & Success (\%)~$\uparrow$ & Agreement (\%) \\
\midrule
\rowcolor{familyshade}\multicolumn{6}{l}{\textit{Qwen3.8-Flash}} \\
w/o Thinking & 0.63\textsubscript{\textpm0.02} & 85.17\textsubscript{\textpm0.76} & 0.40\textsubscript{\textpm0.01} & 12.17\textsubscript{\textpm1.53} & 91.33\textsubscript{\textpm2.25} \\
w/ Thinking & \underline{0.68}\textsubscript{\textpm0.03} & \underline{95.33}\textsubscript{\textpm2.08} & \underline{0.51}\textsubscript{\textpm0.02} & \underline{13.17}\textsubscript{\textpm1.89} & 92.17\textsubscript{\textpm1.26} \\
TMI & \textbf{0.76}\textsubscript{\textpm0.02} & \textbf{96.67}\textsubscript{\textpm1.76} & \textbf{0.73}\textsubscript{\textpm0.01} & \textbf{28.83}\textsubscript{\textpm2.02} & 91.17\textsubscript{\textpm0.76} \\
\midrule
\rowcolor{familyshade}\multicolumn{6}{l}{\textit{DeepSeek-v4-Flash}} \\
w/o Thinking & 0.66\textsubscript{\textpm0.03} & \textbf{93.17}\textsubscript{\textpm1.89} & 0.16\textsubscript{\textpm0.02} & 1.83\textsubscript{\textpm0.29} & 97.17\textsubscript{\textpm2.02} \\
w/ Thinking & \underline{0.67}\textsubscript{\textpm0.02} & \underline{93.00}\textsubscript{\textpm1.00} & \underline{0.38}\textsubscript{\textpm0.002} & \underline{7.17}\textsubscript{\textpm2.47} & 97.17\textsubscript{\textpm0.29} \\
TMI & \textbf{0.69}\textsubscript{\textpm0.02} & \textbf{93.17}\textsubscript{\textpm3.18} & \textbf{0.51}\textsubscript{\textpm0.01} & \textbf{16.00}\textsubscript{\textpm1.73} & 95.33\textsubscript{\textpm0.76} \\
\bottomrule
\end{tabular*}
\caption{P4G and CB buyer results (mean $\pm$ SD across three benchmark repeats). Emphasis follows Table~\ref{tab:social-support-main-results}; agreement rates are descriptive. Percentage changes in the text use unrounded means.}
\label{tab:monetary-main-results}
\endgroup
\end{table*}

\subsection{Experimental Setup}
\label{sec:setup}

We evaluate Qwen3.8-Flash and DeepSeek-v4-Flash under \emph{\mbox{w/o Thinking}}, \emph{\mbox{w/ Thinking}}, and TMI. All conditions share assignments, role information, partners, permitted actions, and generation settings. Each assignment has three rollouts per condition, yielding 15,678 scored main-experiment episodes (Appendix~\ref{app:experimental-details}). Prompts appear in Appendix~\ref{app:interaction-prompts}.

\subsection{Main Results}
\label{sec:main-results}

\noindent\textbf{Overall performance.} TMI consistently achieves the highest mean scores on the four primary outcomes for both models: social goal achievement, emotional-support preference, donation amount, and negotiation margin (Tables~\ref{tab:social-support-main-results} and~\ref{tab:monetary-main-results}). The remaining Sotopia and ESC evaluation metrics appear in Appendix~\ref{app:complete-results}.

\noindent\textbf{Social goals and support quality.} Relative to \emph{\mbox{w/ Thinking}}, TMI raises Sotopia Goal scores from 8.19 to \textbf{8.66} for Qwen and 8.09 to \textbf{8.41} for DeepSeek. Relationship, Knowledge, and Financial/Material Benefits also improve over this baseline, with the highest Overall scores for both models. In ESC-Eval, mean Overall preference scores rise from 1.69 to \textbf{1.90} and 1.72 to \textbf{1.92}, respectively, with gains in Expression, Information, and Skill.

\noindent\textbf{Persuasion and negotiation.} Relative to \emph{\mbox{w/ Thinking}}, TMI raises mean P4G donations by \textbf{12.3\%} for Qwen and \textbf{3.7\%} for DeepSeek, with donation success rates above 93\% for both models. In CB, mean margins rise by \textbf{0.22} for Qwen and \textbf{0.14} for DeepSeek, while success rates more than double. Agreement rates decline slightly but remain above 90\% across both evaluated models.

\section{Interaction Behavior and Reasoning}
\label{sec:analysis}
\label{sec:behavior}

We examine how TMI changes agents' initial proposals, responses to counterpart feedback, and reasoning across turns.

\subsection{Adjusting Proposals and Ending Interactions}
\label{sec:sotopia-behavior}
We examine how agents adjust their actions when initial proposals encounter resistance, focusing on the Tile buyer and weekend-trip inviter. These roles were selected for TMI's Goal gains across both models and baselines (Appendix~\ref{app:sotopia-behavior}).

\begin{table}[t]
\centering\small
\setlength{\tabcolsep}{4pt}\renewcommand{\arraystretch}{1.12}
\begin{tabular*}{\linewidth}{@{\extracolsep{\fill}}lccc@{}}
\toprule
Model & \shortstack{w/o Thinking} & \shortstack{w/ Thinking} & TMI \\
\midrule
\multicolumn{4}{@{}l}{\textit{Tile: opening offer below \$14}} \\
Qwen & 0/15 & 2/15 & \textbf{5/15} \\
DeepSeek & 0/15 & 0/15 & \textbf{5/15} \\
\midrule
\multicolumn{4}{@{}l}{\textit{Trip: commitment inquiry after refusal}} \\
Qwen & 3/15 & 2/15 & \textbf{9/15} \\
DeepSeek & 0/15 & 3/15 & \textbf{5/15} \\
\bottomrule
\end{tabular*}
\caption{Action frequencies in two selected Sotopia roles (15 episodes per model and condition). Definitions and annotation procedures appear in Appendix~\ref{app:sotopia-behavior}.}
\label{tab:sotopia-actions}
\vspace{-0.8em}
\end{table}

\noindent\textbf{Low opening offers and inquiries after refusal.} In Tile, agents seek to buy the item near \$14; TMI more often opens below this target (Table~\ref{tab:sotopia-actions}). All ten low-opening TMI episodes raise offers or accept higher prices, reaching agreements at \$14.50--16 with mean Goal scores of 8.0 for Qwen and 8.2 for DeepSeek. The two low-opening episodes under \emph{\mbox{w/ Thinking}} end in withdrawal with Goal scores of 3. The reviewed reasoning traces in both thinking conditions consider bargaining room; price adjustments secure the item in these TMI episodes. In Trip, TMI more often asks about existing plans after refusal. All four TMI invitations adapted to these plans are rejected. The original weekend invitation fails in all conditions, but TMI obtains more concrete plans for later travel: 13/15 versus 6/15 with \emph{\mbox{w/ Thinking}} for Qwen, and 9/15 versus 6/15 for DeepSeek. TMI obtains more concrete arrangements for later travel after the counterpart declines the original weekend invitation.

\begin{figure}[t]
\centering
\includegraphics[width=\columnwidth]{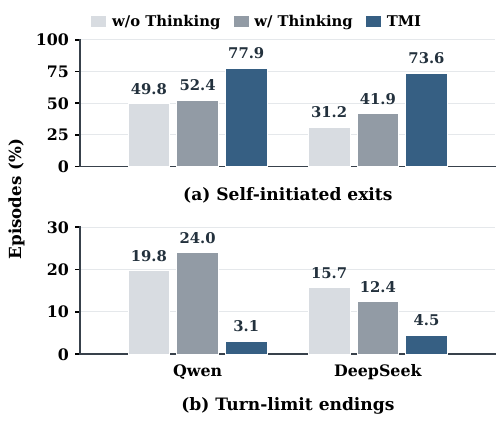}
\caption{Sotopia interaction endings (420 episodes per model and condition): (a) agent-initiated exits; (b) turn-limit terminations. Events are non-exclusive.}
\label{fig:sotopia-closure}
\vspace{-1em}
\end{figure}

\noindent\textbf{Initiative in ending interactions.} Across the full benchmark, TMI more often selects \texttt{leave} to end an interaction (Figure~\ref{fig:sotopia-closure}). Relative to \emph{\mbox{w/ Thinking}}, exits rise from 52.4\% to \textbf{77.9\%} for Qwen and from 41.9\% to \textbf{73.6\%} for DeepSeek, while turn-limit endings fall from 24.0\% to \textbf{3.1\%} and from 12.4\% to \textbf{4.5\%}, respectively. Exit rates increase in 21/28 and 24/28 scene--roles, respectively. TMI more often ends interactions through explicit agent actions, following goal completion or abandonment.

\subsection{Eliciting Context and Adapting Support}
\label{sec:esc-behavior}
We examine how supporters obtain and use information to adapt their responses, focusing on job search, family conflict, and breakup. These scenarios were selected for TMI's Overall gains across both models and baselines (Appendix~\ref{app:esc-behavior}).

\begin{table}[t]
\centering
\begingroup
\small
\setlength{\tabcolsep}{4pt}\renewcommand{\arraystretch}{1.12}
\begin{tabular*}{\linewidth}{@{\extracolsep{\fill}}lccc@{}}
\toprule
Scenario & \shortstack{w/o Thinking} & \shortstack{w/ Thinking} & TMI \\
\midrule
\multicolumn{4}{@{}l}{\textit{Elicit: new information}} \\
Job search & 2/6 & 0/6 & 3/6 \\
Family conflict & 0/6 & 2/6 & 6/6 \\
Breakup & 0/6 & 0/6 & 4/6 \\
\addlinespace[2pt]
Total & 2/18 & 2/18 & \textbf{13/18} \\
\midrule
\multicolumn{4}{@{}l}{\textit{Use: information used in support}} \\
Job search & 1/6 & 0/6 & 2/6 \\
Family conflict & 0/6 & 0/6 & 2/6 \\
Breakup & 0/6 & 0/6 & 3/6 \\
\addlinespace[2pt]
Total & 1/18 & 0/18 & \textbf{7/18} \\
\bottomrule
\end{tabular*}
\endgroup
\caption{Information elicitation and use in three selected ESC-Eval scenarios. Counts indicate qualifying dialogues out of six per scenario and condition.}
\label{tab:esc-information-use}
\vspace{-0.8em}
\end{table}

\begin{figure*}[t]
\centering
\includegraphics[width=\textwidth]{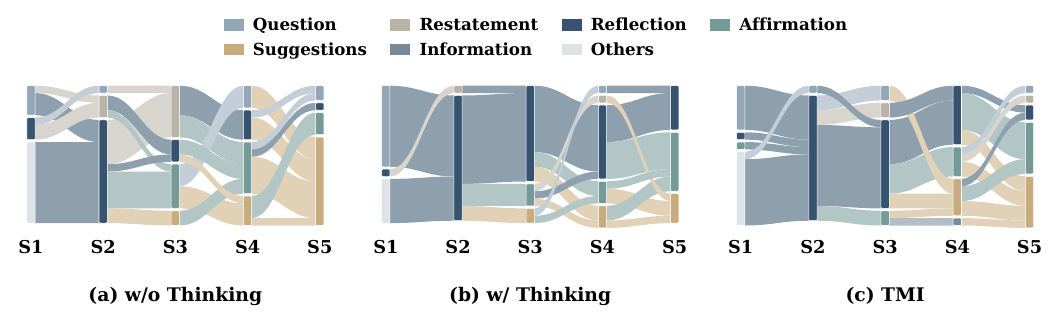}
\caption{Model-selected strategy transitions in three gain-selected ESC-Eval scenarios (Table~\ref{tab:esc-information-use}; 18 dialogues per condition). Node heights and ribbon widths encode dialogue proportions; ribbon colors identify the target strategy.}
\label{fig:esc-strategy-flow}
\end{figure*}

\noindent\textbf{Targeted inquiries and concrete assistance.} TMI elicits new information through targeted inquiries in 13/18 dialogues and uses it in the next reply to provide concrete assistance in 7/18 (Table~\ref{tab:esc-information-use}). This inquiry--information--assistance sequence occurs in 0/18 dialogues with \emph{\mbox{w/ Thinking}} and 1/18 with \emph{\mbox{w/o Thinking}}. In a DeepSeek TMI job-search dialogue, learning that the seeker lacks graphic-design experience leads to portfolio suggestions. When freelance work feels daunting, the supporter narrows the advice to a small personal project.

\noindent\textbf{Reflection and suggestions across turns.} Figure~\ref{fig:esc-strategy-flow} shows the strategies selected over five supporter turns, $S_1$--$S_5$. Both thinking conditions favor reflection at $S_2$, after which TMI more often selects affirmation and suggestions. Across all 331 cards, TMI's reflection share averaged over $S_3$--$S_5$ is \textbf{8.1} percentage points lower than \emph{\mbox{w/ Thinking}}, and its suggestion share at $S_5$ is \textbf{5.4} points higher (Appendix~\ref{app:esc-strategy-flows}). In the three selected scenarios, \emph{\mbox{w/o Thinking}} has a higher suggestion share than TMI at the final supporter turn, $S_5$ (\textbf{66.7\%} vs.\ \textbf{38.9\%} of dialogues in each condition).

\subsection{Opening Offers and Donation Framing}
\label{sec:monetary-behavior}
\begin{figure}[t]
\centering
\includegraphics[width=\columnwidth]{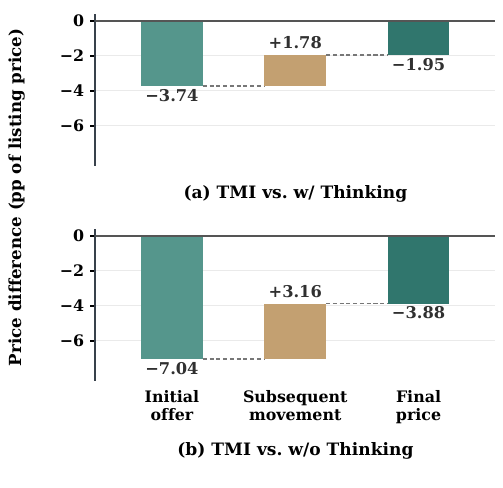}
\caption{CB differences in opening offers, subsequent price movement, and agreed prices (TMI minus baseline), in percentage points of listing price. Values are equal-model means, rounded independently.}
\label{fig:cb-price-decomposition}
\vspace{-0.8em}
\end{figure}

\noindent\textbf{CB: lower opening offers and agreed prices.} We analyze CB assignments where both conditions reach agreement in all three repeats (Appendix~\ref{app:cb-behavior}). Compared with \emph{\mbox{w/ Thinking}}, TMI's opening offers are \textbf{3.74} percentage points of listing price lower. During negotiation, prices rise by \textbf{1.78} points more under TMI, but the agreed prices are still \textbf{1.95} points lower (Figure~\ref{fig:cb-price-decomposition}). Both models show this pattern against both baselines. In the reviewed traces, models in both thinking conditions consider bargaining room but choose different opening prices.

\noindent\textbf{P4G: fewer low opening amounts.} We focus on personas where \emph{\mbox{w/ Thinking}} mentions a positive donation amount below \$1 in an opening reply, but TMI makes no such opening mention. In this group, TMI's mean donations are \textbf{\$0.21} higher for Qwen (88 personas) and \textbf{\$0.16} higher for DeepSeek (49). When neither condition mentions low opening amounts, TMI's mean donations are similar to or slightly below the baseline's (Appendix~\ref{app:p4g-behavior}). Each episode ends at the first valid donation decision. An amount presented as a small starting point can therefore become the final donation.

\begin{figure*}[t]
\centering
\includegraphics[width=\textwidth]{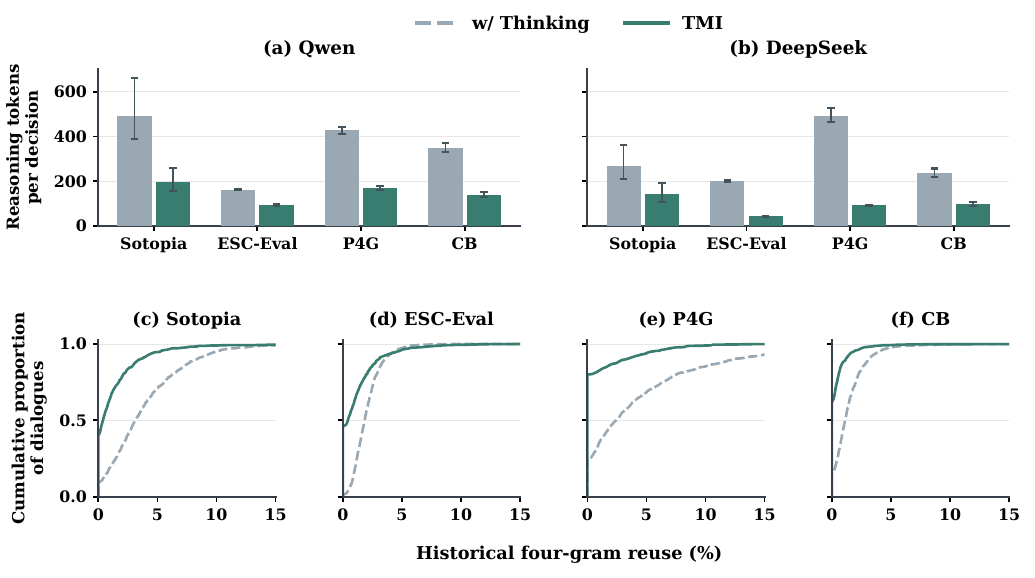}
\caption{Reasoning tokens per decision (a--b; means with 95\% cluster-bootstrap confidence intervals) and cross-turn four-gram reuse (c--f; dialogue-level ECDFs).}
\label{fig:reasoning-length-repetition}
\end{figure*}

\subsection{More Concise Reasoning Across Turns}
\label{sec:reasoning-behavior}
\noindent\textbf{Shorter reasoning per decision.} Across all eight model--environment pairs, TMI reduces mean reasoning tokens per decision by \textbf{42.0--81.5\%} relative to \emph{\mbox{w/ Thinking}} (Figure~\mbox{\ref{fig:reasoning-length-repetition}a--b}). ESC-Eval fixes the number of supporter turns at five in both conditions. In P4G and CB, TMI makes slightly more decisions per dialogue on average than the thinking baseline. Yet TMI generates fewer reasoning tokens per dialogue in all three environments. The per-decision reduction holds when comparing the first three decisions of paired dialogues that reach three decisions in both conditions.

\noindent\textbf{Reduced reuse of earlier reasoning.} We measure cross-turn phrase reuse as the proportion of content-word four-gram occurrences in current reasoning that also appear in earlier reasoning from the same agent. TMI's mean reuse is \textbf{0.76--5.95} percentage points lower than \emph{\mbox{w/ Thinking}} across all eight model--environment pairs. The ECDFs also show a larger share of dialogues with low reuse under TMI (Figure~\mbox{\ref{fig:reasoning-length-repetition}c--f}). When we match the lengths of both current and historical reasoning across conditions, TMI still has lower reuse in seven of the eight model--environment pairs. TMI thus preserves reasoning history across turns while generating shorter traces with less reuse of earlier phrasing. Further details are provided in Appendix~\ref{app:reasoning-behavior}.

\section{Ablation Studies}
\label{sec:ablation}
\label{sec:preserved-ablation}
To test whether reasoning retention accounts for TMI's gains, we use Qwen3.8-Flash, which supports reasoning retention in conversational requests. DeepSeek-v4-Flash retains prior reasoning only in tool-calling contexts \citep{xu2026deepseek}.

\begin{table}[t]
\centering\small
\setlength{\tabcolsep}{2pt}\renewcommand{\arraystretch}{1.18}
\begin{tabular*}{\columnwidth}{@{\extracolsep{\fill}}lccc@{}}
\toprule
Outcome & \shortstack{w/ Thinking} & Preserved & TMI \\
\midrule
Sotopia Goal & \underline{8.19}\textsubscript{\textpm0.12} & 8.01\textsubscript{\textpm0.23} & \textbf{8.66}\textsubscript{\textpm0.04} \\
ESC Overall & \underline{1.69}\textsubscript{\textpm0.02} & 1.61\textsubscript{\textpm0.003} & \textbf{1.90}\textsubscript{\textpm0.003} \\
P4G Donation & \underline{0.68}\textsubscript{\textpm0.03} & 0.67\textsubscript{\textpm0.03} & \textbf{0.76}\textsubscript{\textpm0.02} \\
CB Margin & 0.51\textsubscript{\textpm0.02} & \underline{0.57}\textsubscript{\textpm0.02} & \textbf{0.73}\textsubscript{\textpm0.01} \\
\bottomrule
\end{tabular*}
\caption{Qwen reasoning-preservation ablation (mean $\pm$ SD). CB reports margin; P4G reports dollars.}
\label{tab:reasoning-preservation}
\vspace{-0.8em}
\end{table}

\noindent\textbf{Task outcomes.} The \emph{Preserved} condition retains prior reasoning in assistant--user message exchanges, using the same assignments and three rollouts per assignment as the other conditions (2,613 additional episodes). Relative to \emph{\mbox{w/ Thinking}}, Preserved has lower mean Sotopia Goal and ESC Overall scores and lower mean P4G donations. Its mean CB margin increases from 0.51 to 0.57 but remains below TMI's \textbf{0.73}. TMI has higher means than Preserved on all four outcomes (Table~\ref{tab:reasoning-preservation}).

\noindent\textbf{Reasoning length and phrase reuse.} Relative to \emph{\mbox{w/ Thinking}}, Preserved reduces mean reasoning tokens per decision in Sotopia and ESC but increases them in P4G and CB. TMI generates \textbf{35.0--62.2\%} fewer reasoning tokens per decision than Preserved across all four environments. On reasoning prefixes matched across all three conditions, Preserved has lower mean historical n-gram reuse than \emph{\mbox{w/ Thinking}}. TMI has lower mean reuse than Preserved in all four environments for $n=2,3,4$. Detailed comparisons appear in Appendix~\ref{app:reasoning-preservation-control}.
\section{Discussion}
\label{sec:discussion}
\noindent\textbf{Goal-directed action across exchanges.} The behavioral patterns suggest that TMI helps models translate dialogue-level goals into action choices across successive exchanges. In CB and P4G, opening offers and donation framing are associated with final monetary outcomes. In the reviewed Sotopia and ESC-Eval trajectories, information obtained from the counterpart guides subsequent adjustments to proposals and support. Together, these findings highlight two complementary aspects of goal pursuit: establishing conditions for later progress and adapting actions as new information becomes available. This provides a behavioral interpretation of how native agency in tool use can extend to goal-directed dialogue.

\Needspace{3\baselineskip}
\noindent\textbf{Unified execution and agentic training.} TMI's task gains are accompanied by shorter reasoning traces with less reuse of earlier phrasing, despite retaining reasoning history. TMI also outperforms the Qwen reasoning-retention control on all four primary outcomes and both reasoning measures, highlighting the broader organization of reasoning, actions, and observations. Beyond dialogue-only environments, the same interface could combine real tool operations and dialogue actions within one continuing execution loop, allowing both to advance a shared goal across turns. Its compatibility with native agentic execution further suggests a route for incorporating multi-turn dialogue into agentic post-training, where outcomes across exchanges could guide joint learning of tool use and dialogue policies.

\section{Conclusion}
We investigated TMI, a training-free protocol for goal-directed multi-turn dialogue through native tool execution. Across four environments and two models, TMI improves primary outcomes and generates shorter reasoning with less cross-turn phrase reuse. For Qwen, TMI also outperforms a reasoning-preservation control on task outcomes and reasoning measures. These findings show that interaction organization shapes models' goal pursuit across dialogue turns.

\section*{Limitations}

\noindent\textbf{Model coverage.} Our evaluation covers two models interacting with simulated counterparts. Extending the comparison to other model families, tool interfaces, and human partners would establish how broadly the findings apply.

\noindent\textbf{Behavioral interpretation.} Our analyses characterize associations between observable actions, natural-language reasoning traces, and task outcomes. The Sotopia and ESC case analyses focus on roles and scenarios selected for TMI's gains. These cases describe behavior within those cohorts; identifying causal mechanisms would require targeted interventions or analyses of internal model representations.

\section*{Ethical Considerations}

This study examines goal-directed dialogue using existing benchmarks and simulated partners. Reported donations and negotiation agreements are benchmark outcomes and involve no real financial transactions. Improved goal attainment does not by itself establish that an interaction benefits all participants. Applications of sustained dialogue agency should therefore respect participants' autonomy and informed choices, particularly in persuasion and emotional support.

\bibliography{references}
\clearpage
\appendix
\raggedbottom
\makeatletter
\setlength{\@fptop}{0pt}
\setlength{\@fpsep}{18pt}
\setlength{\@fpbot}{0pt plus 1fil}
\setlength{\@dblfptop}{0pt}
\setlength{\@dblfpsep}{18pt}
\setlength{\@dblfpbot}{0pt plus 1fil}
\makeatother
\suppressfloats[t]
\section{Experimental Details}
\label{app:experimental-details}
\noindent\textbf{Models and generation.} Both models use temperature $1.0$ and an 8,192-token output limit. Qwen uses top-$p$ $0.95$ and top-$k$ $20$; DeepSeek uses top-$p$ $1.0$. These settings are fixed across conditions. For counterparts, Sotopia uses \texttt{gpt-4o-2024-08-06} at temperature $0.7$. ESC-Eval uses the official ESC-Role seeker with temperature $0.7$, top-$p$ $0.8$, top-$k$ $20$, and a 512-token output limit. P4G and CB use GPT-6-Luna with an 8,192-token output limit.

\noindent\textbf{Assignments and visibility.} Sotopia includes 70 profile-pair assignments across 14 Hard scenarios, with each participant role evaluated separately. ESC uses all 331 English high-quality cards \citep{zhao-etal-2024-esc}, providing only the base role description to the seeker. P4G uses METRO's 200 evaluation personas, with ten per personality--decision-style combination; personas are visible only to counterparts. CB uses METRO's 200 development assignments from 85 listings, with the buyer speaking first. In ESC, P4G, and CB, counterparts receive only public responses; reasoning and strategy labels remain private. Sotopia follows its participant-specific action visibility rules.

\noindent\textbf{Scoring and termination.} Each public Sotopia dialogue is scored once by GPT-4o at temperature $0$ and top-$p$ $1$. ESC uses the official English scoring prompts with InternLM2-chat-7b and seven ESC-RANK LoRA adapters, at temperature $0$ and a 32-token output limit. P4G ends at the first valid donation decision, which specifies the donation amount. CB agreement requires consecutive matching price confirmations; a no-deal declaration or the action limit ends negotiation without agreement.

\section{Complete Results}
\label{app:complete-results}
Tables~\ref{tab:sotopia-full-results} and~\ref{tab:esc-full-results} supplement Table~\ref{tab:social-support-main-results} with all Sotopia and ESC metrics for both models and all three conditions. P4G and CB buyer metrics are fully reported in Table~\ref{tab:monetary-main-results}. Each assignment has three rollouts per condition. Tables report the mean and sample SD across the three complete-benchmark repeat means.

\begin{table}[!tbp]
\centering\small
\setlength{\tabcolsep}{3pt}
\begin{tabular*}{\linewidth}{@{\extracolsep{\fill}}lcc@{}}
\toprule
Group & Qwen & DeepSeek \\
\midrule
\multicolumn{3}{l}{\textit{P4G: donation (\$)}} \\
Analytical & $+0.108\pm0.047$ & $-0.015\pm0.066$ \\
Behavioral & $+0.012\pm0.075$ & $+0.017\pm0.025$ \\
Conceptual & $+0.130\pm0.066$ & $+0.041\pm0.004$ \\
Directive & $+0.083\pm0.024$ & $+0.055\pm0.017$ \\
\midrule
\multicolumn{3}{l}{\textit{CB: buyer margin}} \\
Bike & $+0.173\pm0.030$ & $+0.140\pm0.027$ \\
Car & $+0.220\pm0.150$ & $+0.118\pm0.029$ \\
Electronics & $+0.222\pm0.067$ & $+0.164\pm0.069$ \\
Furniture & $+0.238\pm0.088$ & $+0.160\pm0.041$ \\
Housing & $+0.176\pm0.069$ & $+0.094\pm0.024$ \\
Phone & $+0.247\pm0.063$ & $+0.119\pm0.035$ \\
\bottomrule
\end{tabular*}
\caption{Subgroup gains over w/ Thinking (mean $\pm$ SD across three repeats). SD describes variability across repeat means.}
\label{tab:subgroup-gains}
\end{table}

\subsection{Subgroup Results}
\label{app:subgroup-results}
Table~\ref{tab:subgroup-gains} reports TMI-minus-\emph{\mbox{w/ Thinking}} differences in P4G donation amount by decision style (50 personas per style) and CB negotiation margin by item category. For each subgroup, we subtract the baseline mean from the TMI mean within each repeat and report the mean and sample SD of the three differences.

\begin{table*}[!tp]
\centering\small
\setlength{\tabcolsep}{1.5pt}\renewcommand{\arraystretch}{1.18}
\begin{tabular*}{\textwidth}{@{\extracolsep{\fill}}lcccccccc@{}}
\toprule
Setting & Goal~$\uparrow$ & Bel.~$\uparrow$ & Rel.~$\uparrow$ & Know.~$\uparrow$ & Secret~$\uparrow$ & Rules~$\uparrow$ & Fin.~$\uparrow$ & Overall~$\uparrow$ \\
\midrule
\rowcolor{familyshade}\multicolumn{9}{l}{\textit{Qwen3.8-Flash}} \\
w/o Thinking & 7.92\textsubscript{\textpm0.15} & \underline{9.05}\textsubscript{\textpm0.03} & 1.85\textsubscript{\textpm0.05} & 4.73\textsubscript{\textpm0.08} & \textminus{}0.04\textsubscript{\textpm0.01} & \textminus{}0.25\textsubscript{\textpm0.04} & 1.40\textsubscript{\textpm0.06} & 3.52\textsubscript{\textpm0.01} \\
w/ Thinking & \underline{8.19}\textsubscript{\textpm0.12} & 9.03\textsubscript{\textpm0.02} & \underline{2.14}\textsubscript{\textpm0.13} & \underline{5.15}\textsubscript{\textpm0.30} & \textbf{0.00}\textsubscript{\textpm0.00} & \textbf{\textminus{}0.08}\textsubscript{\textpm0.06} & \underline{1.49}\textsubscript{\textpm0.04} & \underline{3.70}\textsubscript{\textpm0.05} \\
TMI & \textbf{8.66}\textsubscript{\textpm0.04} & \textbf{9.09}\textsubscript{\textpm0.004} & \textbf{2.71}\textsubscript{\textpm0.10} & \textbf{5.43}\textsubscript{\textpm0.32} & \underline{\textminus{}0.01}\textsubscript{\textpm0.02} & \underline{\textminus{}0.13}\textsubscript{\textpm0.03} & \textbf{1.65}\textsubscript{\textpm0.02} & \textbf{3.91}\textsubscript{\textpm0.06} \\
\midrule
\rowcolor{familyshade}\multicolumn{9}{l}{\textit{DeepSeek-v4-Flash}} \\
w/o Thinking & 8.02\textsubscript{\textpm0.10} & 9.05\textsubscript{\textpm0.02} & \textbf{2.48}\textsubscript{\textpm0.03} & \underline{5.33}\textsubscript{\textpm0.27} & \textminus{}0.06\textsubscript{\textpm0.02} & \underline{\textminus{}0.26}\textsubscript{\textpm0.05} & \underline{1.55}\textsubscript{\textpm0.08} & \underline{3.73}\textsubscript{\textpm0.04} \\
w/ Thinking & \underline{8.09}\textsubscript{\textpm0.03} & \underline{9.06}\textsubscript{\textpm0.01} & 2.28\textsubscript{\textpm0.07} & 5.27\textsubscript{\textpm0.12} & \textbf{\textminus{}0.03}\textsubscript{\textpm0.03} & \textminus{}0.26\textsubscript{\textpm0.05} & 1.32\textsubscript{\textpm0.04} & 3.67\textsubscript{\textpm0.01} \\
TMI & \textbf{8.41}\textsubscript{\textpm0.13} & \textbf{9.07}\textsubscript{\textpm0.01} & \underline{2.42}\textsubscript{\textpm0.06} & \textbf{5.34}\textsubscript{\textpm0.19} & \underline{\textminus{}0.05}\textsubscript{\textpm0.05} & \textbf{\textminus{}0.24}\textsubscript{\textpm0.05} & \textbf{1.62}\textsubscript{\textpm0.03} & \textbf{3.80}\textsubscript{\textpm0.05} \\
\bottomrule
\end{tabular*}
\caption{Sotopia-Hard results (mean $\pm$ SD). Overall averages the seven raw dimensions.}
\label{tab:sotopia-full-results}
\end{table*}

\begin{table*}[!tp]
\centering\small
\setlength{\tabcolsep}{1.5pt}\renewcommand{\arraystretch}{1.18}
\begin{tabular*}{\textwidth}{@{\extracolsep{\fill}}lccccccc@{}}
\toprule
Setting & Fluency~$\uparrow$ & Expression~$\uparrow$ & Empathy~$\uparrow$ & Information~$\uparrow$ & Humanoid~$\uparrow$ & Skill~$\uparrow$ & Overall~$\uparrow$ \\
\midrule
\rowcolor{familyshade}\multicolumn{8}{l}{\textit{Qwen3.8-Flash}} \\
w/o Thinking & 3.00\textsubscript{\textpm0.00} & 2.84\textsubscript{\textpm0.02} & 3.00\textsubscript{\textpm0.00} & 2.22\textsubscript{\textpm0.04} & \underline{2.64}\textsubscript{\textpm0.02} & 2.89\textsubscript{\textpm0.03} & 1.47\textsubscript{\textpm0.02} \\
w/ Thinking & 3.00\textsubscript{\textpm0.00} & \underline{2.93}\textsubscript{\textpm0.03} & 3.00\textsubscript{\textpm0.00} & \underline{2.34}\textsubscript{\textpm0.01} & \textbf{2.70}\textsubscript{\textpm0.03} & \underline{2.95}\textsubscript{\textpm0.02} & \underline{1.69}\textsubscript{\textpm0.02} \\
TMI & 3.00\textsubscript{\textpm0.00} & \textbf{2.96}\textsubscript{\textpm0.02} & 3.00\textsubscript{\textpm0.00} & \textbf{2.45}\textsubscript{\textpm0.02} & 2.63\textsubscript{\textpm0.02} & \textbf{2.98}\textsubscript{\textpm0.01} & \textbf{1.90}\textsubscript{\textpm0.003} \\
\midrule
\rowcolor{familyshade}\multicolumn{8}{l}{\textit{DeepSeek-v4-Flash}} \\
w/o Thinking & 3.00\textsubscript{\textpm0.00} & 2.88\textsubscript{\textpm0.01} & 3.00\textsubscript{\textpm0.00} & 2.14\textsubscript{\textpm0.02} & \underline{2.73}\textsubscript{\textpm0.01} & 2.91\textsubscript{\textpm0.02} & 1.50\textsubscript{\textpm0.02} \\
w/ Thinking & 3.00\textsubscript{\textpm0.00} & \underline{2.94}\textsubscript{\textpm0.02} & 3.00\textsubscript{\textpm0.00} & \underline{2.22}\textsubscript{\textpm0.002} & \textbf{2.75}\textsubscript{\textpm0.03} & \underline{2.96}\textsubscript{\textpm0.01} & \underline{1.72}\textsubscript{\textpm0.02} \\
TMI & 3.00\textsubscript{\textpm0.00} & \textbf{2.96}\textsubscript{\textpm0.01} & 3.00\textsubscript{\textpm0.00} & \textbf{2.25}\textsubscript{\textpm0.03} & 2.69\textsubscript{\textpm0.03} & \textbf{2.99}\textsubscript{\textpm0.003} & \textbf{1.92}\textsubscript{\textpm0.01} \\
\bottomrule
\end{tabular*}
\caption{ESC-Eval scores (0--4; mean $\pm$ SD). Overall is rated separately.}
\label{tab:esc-full-results}
\end{table*}

\section{Behavioral and Reasoning Results}
\label{app:behavioral-results}
\label{app:supplementary-analysis}
Two doctoral students reviewed and corrected the Sotopia and ESC-Eval annotations using public dialogue and available reasoning. The following subsections provide annotation criteria, extraction rules, and detailed statistics for the analyses in Sections~\ref{sec:analysis} and~\ref{sec:ablation}.

\subsection{Sotopia}
\label{app:sotopia-behavior}
\noindent\textbf{Case selection.} For each of the 28 scene--role objectives, we compare TMI with the thinking and non-thinking baselines separately for Qwen and DeepSeek, yielding four Goal-score gains. The smallest of the four gains serves as the selection score. The two objectives with the highest selection scores are the Tile buyer and weekend-trip inviter, each covering 90 episodes. Goal scores are averaged across repeats within each profile, with equal profile weights.

\noindent\textbf{Annotation criteria.} The annotations support the action counts in Table~\ref{tab:sotopia-actions} and the trajectory outcomes in Section~\ref{sec:sotopia-behavior}. In Tile, a low opening is the buyer's first stated purchase offer or budget below \$14; references to listing or market prices are excluded. For these low-opening episodes, we record whether the participants agree on terms for the original item, the agreed price, and the benchmark Goal score. In Trip, inquiry is recorded when the first action after the initial refusal asks about the content, type, or duration of the counterpart's existing plans. Generic requests for flexibility are excluded. We also record whether the next invitation accommodates the disclosed constraint and whether it is accepted. We count a later travel arrangement when both participants agree on a specific next step in planning the trip or a later time to travel. General expressions of interest are excluded.

\begin{table}[!tbp]
\centering\small
\setlength{\tabcolsep}{2pt}\renewcommand{\arraystretch}{1.10}
\begin{tabular*}{\linewidth}{@{\extracolsep{\fill}}llccc@{}}
\toprule
\multicolumn{5}{l}{\textit{Tile: low-opening episodes}} \\
Model & Condition & \shortstack{Low\\openings} & Agreements & \shortstack{Mean\\Goal} \\
\midrule
Qwen & w/o Thinking & 0/15 & -- & -- \\
Qwen & w/ Thinking & 2/15 & 0/2 & 3.0 \\
Qwen & TMI & 5/15 & 5/5 & 8.0 \\
DeepSeek & w/o Thinking & 0/15 & -- & -- \\
DeepSeek & w/ Thinking & 0/15 & -- & -- \\
DeepSeek & TMI & 5/15 & 5/5 & 8.2 \\
\end{tabular*}

\begin{tabular*}{\linewidth}{@{\extracolsep{\fill}}llcc@{}}
\toprule
\multicolumn{4}{l}{\textit{Trip: inquiries and later arrangements}} \\
Model & Condition & Inquiries & \shortstack{Later\\arrangements} \\
\midrule
Qwen & w/o Thinking & 3/15 & 1/15 \\
Qwen & w/ Thinking & 2/15 & 6/15 \\
Qwen & TMI & 9/15 & 13/15 \\
DeepSeek & w/o Thinking & 0/15 & 5/15 \\
DeepSeek & w/ Thinking & 3/15 & 6/15 \\
DeepSeek & TMI & 5/15 & 9/15 \\
\bottomrule
\end{tabular*}
\caption{Statistics supporting the Sotopia analysis (15 episodes per model and condition). Tile agreement counts and mean Goal scores cover only low-opening episodes; -- indicates no such episode.}
\label{tab:sotopia-role-outcomes}
\end{table}

\noindent\textbf{Interaction endings.} Statistics cover all 2,520 episodes. Agent-initiated exits are identified by the evaluated agent's \texttt{leave} action, and turn-limit terminations by the logged ending reason. The two events are counted separately and may co-occur.

\subsection{ESC-Eval}
\label{app:esc-behavior}
\label{app:esc-strategy-flows}
\noindent\textbf{Case selection.} For each of the 331 cards, we average Overall scores over three repeats and compare TMI with the thinking and non-thinking baselines separately for Qwen and DeepSeek. The smallest of these four gains serves as the selection score. We select the three cards with the highest selection scores, breaking ties by mean gain: job search (ID 32381), family conflict (ID 32084), and breakup (ID 32186). Annotation covers all 54 dialogues across two models, three conditions, and three repeats.

\noindent\textbf{Annotation criteria.} \textit{Elicit} counts dialogues in which a supporter inquiry at $S_2$--$S_4$ asks for a detail not established earlier, and the next seeker reply provides it. These details may concern goals, prior attempts, or constraints. Opening questions, generic requests to elaborate, and permission checks are excluded. \textit{Use} counts dialogues in which the immediately following supporter reply uses that information to offer a concrete suggestion or example wording, or to reframe the seeker's situation. Acknowledgment, restatement, or further inquiry alone is insufficient. Each dialogue is counted once per measure if it contains a qualifying sequence; every Use dialogue is also counted under Elicit.

\begin{table}[!tbp]
\centering\small
\setlength{\tabcolsep}{4pt}\renewcommand{\arraystretch}{1.08}
\begin{tabular*}{\linewidth}{@{\extracolsep{\fill}}llcc@{}}
\toprule
Model & Condition & Elicit & Use \\
\midrule
Qwen & w/o Thinking & 0/9 & 0/9 \\
Qwen & w/ Thinking & 2/9 & 0/9 \\
Qwen & TMI & 6/9 & 5/9 \\
\midrule
DeepSeek & w/o Thinking & 2/9 & 1/9 \\
DeepSeek & w/ Thinking & 0/9 & 0/9 \\
DeepSeek & TMI & 7/9 & 2/9 \\
\bottomrule
\end{tabular*}
\caption{Model breakdown of Table~\ref{tab:esc-information-use} (nine dialogues per model and condition). Counts sum to the totals in that table.}
\label{tab:esc-process-summary}
\end{table}

\noindent\textbf{Strategy statistics.} Strategy labels are extracted from model outputs at supporter turns $S_1$--$S_5$. At each turn, a strategy's share is the proportion of dialogues carrying that label, with equal model and repeat weights. Reflection averages the shares at $S_3$--$S_5$; suggestions use the share at $S_5$. Table~\ref{tab:esc-flow-sensitivity} reports TMI-minus-thinking differences for the three scenarios in Figure~\ref{fig:esc-strategy-flow} and all 331 cards.

\begin{table}[!tbp]
\centering\small
\setlength{\tabcolsep}{3pt}\renewcommand{\arraystretch}{1.08}
\begin{tabular*}{\linewidth}{@{\extracolsep{\fill}}lrrrr@{}}
\toprule
Scope & Cards & $n$ & $\Delta$Reflect. & $\Delta$Suggest. \\
\midrule
Selected scenarios & 3 & 18 & $-13.0$ & $+16.7$ \\
All scenarios & 331 & 1,986 & $-8.1$ & $+5.4$ \\
\bottomrule
\end{tabular*}
\caption{Strategy shares, TMI minus w/ Thinking (percentage points). $n$: dialogues per condition. Reflection averages $S_3$--$S_5$; suggestions are measured at $S_5$.}
\label{tab:esc-flow-sensitivity}
\end{table}

\setlength{\dblfloatsep}{8pt}
\setlength{\dbltextfloatsep}{12pt}
\subsection{CB Price Paths}
\label{app:cb-behavior}
\noindent\textbf{Extraction and eligibility.} Opening offers are extracted from explicit buyer amounts in all 3,600 main episodes; agreed prices come from environment outcomes. Multiple proposed amounts, unclear price ownership, or a possibly missed earlier offer make the opening unavailable. The parser was calibrated on 24 annotated episodes. Each model--baseline comparison retains assignments with recognized openings and agreements in all three repeats under both conditions (Table~\ref{tab:cb-price-paths}). Qualitative review covers 27 public dialogues from three model--assignment units and four opening reasoning traces.

\noindent\textbf{Decomposition.} For listing price $L$, let $p_0=100P_0/L$ and $p_f=100P_f/L$ denote normalized opening and agreed prices. We average three repeats within assignments and compute TMI-minus-baseline differences, weighting assignments equally within models:
\begin{equation}
\Delta p_f=\Delta p_0+\Delta(p_f-p_0).
\label{eq:price-path}
\end{equation}
Movement includes revised offers and acceptance of the seller's price. Figure~\ref{fig:cb-price-decomposition} weights the two model means equally.

\begin{table}[!htbp]
\centering\small
\setlength{\tabcolsep}{3pt}
\begin{tabular*}{\linewidth}{@{\extracolsep{\fill}}lrrrr@{}}
\toprule
Model & $n$ & Initial & Movement & Final \\
\midrule
\multicolumn{5}{l}{\textit{TMI minus w/ Thinking}} \\
Qwen & 30 & $-5.62$ & $+3.28$ & $-2.35$ \\
DeepSeek & 101 & $-1.85$ & $+0.29$ & $-1.56$ \\
Equal-model mean & -- & $-3.74$ & $+1.78$ & $-1.95$ \\
\midrule
\multicolumn{5}{l}{\textit{TMI minus w/o Thinking}} \\
Qwen & 7 & $-9.33$ & $+5.31$ & $-4.01$ \\
DeepSeek & 74 & $-4.76$ & $+1.01$ & $-3.75$ \\
Equal-model mean & -- & $-7.04$ & $+3.16$ & $-3.88$ \\
\bottomrule
\end{tabular*}
\caption{CB price-path differences in percentage points of listing price. $n$ counts eligible assignments, each contributing three rollouts per condition. Values are rounded independently.}
\label{tab:cb-price-paths}
\end{table}

\begin{table}[!tbp]
\centering\small
\setlength{\tabcolsep}{3pt}\renewcommand{\arraystretch}{1.08}
\begin{tabular*}{\linewidth}{@{\extracolsep{\fill}}lrrrr@{}}
\toprule
Model & Personas & Thinking & TMI & Difference \\
\midrule
\multicolumn{5}{l}{\textit{Low amount mentioned only under w/ Thinking}} \\
Qwen & 88 & 0.57 & 0.78 & $+0.21$ \\
DeepSeek & 49 & 0.53 & 0.69 & $+0.16$ \\
\midrule
\multicolumn{5}{l}{\textit{No low amount mentioned under either condition}} \\
Qwen & 96 & 0.79 & 0.78 & $-0.01$ \\
DeepSeek & 146 & 0.71 & 0.70 & $-0.02$ \\
\bottomrule
\end{tabular*}
\caption{P4G donation means for the groups in Section~\ref{sec:monetary-behavior} (dollars). Differences are TMI minus w/ Thinking; values are rounded independently.}
\label{tab:p4g-opening-groups}
\end{table}

\subsection{P4G Opening Amounts}
\label{app:p4g-behavior}
Opening amounts are extracted from the first public persuader reply in all 3,600 main episodes. A low amount is an explicit positive donation below \$1, excluding the \$2 payment and funds retained by the donor; it need not be a direct request. A persona is classified as having a low opening if any of its three repeats contains one. Table~\ref{tab:p4g-opening-groups} reports the two groups discussed in Section~\ref{sec:monetary-behavior}. Grouping and donation means use the same three repeats; means are averaged within personas, then across personas with equal weights.

\begin{table*}[!tp]
\centering\small
\setlength{\tabcolsep}{3pt}
\begin{tabular*}{\textwidth}{@{\extracolsep{\fill}}llrrrrrrr@{}}
\toprule
& & \multicolumn{2}{c}{Tokens / decision} & Reduction & \multicolumn{2}{c}{Tokens / dialogue} & \multicolumn{2}{c}{Decisions / dialogue} \\
\cmidrule(lr){3-4}\cmidrule(lr){6-7}\cmidrule(l){8-9}
Environment & Model & Thinking & TMI & (\%) & Thinking & TMI & Thinking & TMI \\
\midrule
Sotopia & Qwen & 490.4 & 196.7 & 59.9 & 3351.5 & 1143.4 & 6.86 & 5.80 \\
 & DeepSeek & 269.1 & 142.0 & 47.2 & 1606.9 & 887.4 & 6.35 & 6.14 \\
\addlinespace[2pt]
ESC-Eval & Qwen & 162.2 & 94.0 & 42.0 & 810.8 & 469.9 & 5.00 & 5.00 \\
 & DeepSeek & 200.9 & 43.1 & 78.6 & 1004.6 & 215.3 & 5.00 & 5.00 \\
\addlinespace[2pt]
P4G & Qwen & 426.4 & 168.8 & 60.4 & 979.0 & 424.9 & 2.30 & 2.56 \\
 & DeepSeek & 493.3 & 91.2 & 81.5 & 1146.6 & 249.2 & 2.44 & 2.72 \\
\addlinespace[2pt]
CB & Qwen & 349.6 & 139.8 & 60.0 & 2339.3 & 983.7 & 6.49 & 6.59 \\
 & DeepSeek & 236.5 & 97.2 & 58.9 & 1564.7 & 718.5 & 6.26 & 6.85 \\
\bottomrule
\end{tabular*}
\caption{Reasoning usage, with assignments weighted equally. Each model and condition has 420 dialogues in Sotopia, 993 in ESC-Eval, and 600 in P4G and CB.}
\label{tab:reasoning-usage}
\end{table*}

\begin{table*}[!tp]
\centering\small
\setlength{\tabcolsep}{3pt}
\begin{tabular*}{\textwidth}{@{\extracolsep{\fill}}llrrrrrr@{}}
\toprule
& & \multicolumn{2}{c}{Paired sample} & \multicolumn{2}{c}{Full-text mean (\%)} & \multicolumn{2}{c}{After length matching (pp)} \\
\cmidrule(lr){3-4}\cmidrule(lr){5-6}\cmidrule(l){7-8}
Environment & Model & Dialogues & Assignments & Thinking & TMI & TMI $-$ Thinking & 95\% CI \\
\midrule
Sotopia & Qwen & 413 & 138 & 4.62 & 1.38 & $-8.49$ & $[-10.12, -6.99]$ \\
 & DeepSeek & 419 & 140 & 3.11 & 1.17 & $-3.26$ & $[-4.30, -2.26]$ \\
\addlinespace[2pt]
ESC-Eval & Qwen & 988 & 331 & 2.07 & 1.05 & $-1.18$ & $[-1.31, -1.05]$ \\
 & DeepSeek & 933 & 331 & 2.01 & 1.17 & $+0.04$ & $[-0.14, +0.21]$ \\
\addlinespace[2pt]
P4G & Qwen & 551 & 197 & 2.42 & 0.88 & $-1.47$ & $[-1.80, -1.15]$ \\
 & DeepSeek & 557 & 199 & 6.59 & 0.64 & $-1.38$ & $[-1.79, -0.97]$ \\
\addlinespace[2pt]
CB & Qwen & 590 & 200 & 1.69 & 0.33 & $-1.61$ & $[-1.82, -1.41]$ \\
 & DeepSeek & 551 & 200 & 1.17 & 0.41 & $-1.14$ & $[-1.45, -0.87]$ \\
\bottomrule
\end{tabular*}
\caption{Four-gram reuse with equal assignment weights. Paired counts apply to each condition. Length-matched differences use recomputed reuse rates on truncated texts.}
\label{tab:reasoning-reuse}
\end{table*}

\begin{table}[!tbp]
\centering\small
\setlength{\tabcolsep}{2pt}
\begin{tabular*}{\linewidth}{@{\extracolsep{\fill}}lrrrr@{}}
\toprule
Environment & Pairs & Thinking & TMI & \shortstack{Reduction\\(\%)} \\
\midrule
\multicolumn{5}{l}{\textit{Qwen}} \\
Sotopia & 406 & 459.0 & 217.7 & 52.6 \\
ESC-Eval & 993 & 167.5 & 96.8 & 42.2 \\
P4G & 132 & 430.7 & 159.8 & 62.9 \\
CB & 588 & 303.3 & 99.9 & 67.1 \\
\midrule
\multicolumn{5}{l}{\textit{DeepSeek}} \\
Sotopia & 411 & 296.1 & 124.7 & 57.9 \\
ESC-Eval & 993 & 199.7 & 45.6 & 77.2 \\
P4G & 184 & 399.4 & 92.9 & 76.7 \\
CB & 592 & 225.1 & 63.6 & 71.8 \\
\bottomrule
\end{tabular*}
\caption{Mean reasoning tokens per decision over the first three decisions of paired dialogues reaching three decisions in both conditions.}
\label{tab:reasoning-first-three}
\end{table}

\subsection{Reasoning Length and Phrase Reuse}
\label{app:reasoning-behavior}
\noindent\textbf{Reasoning-token statistics.} We count API-reported reasoning tokens from evaluated-model calls only. For each dialogue, mean tokens per decision equal total reasoning tokens divided by the number of decisions. We average the three rollouts within each assignment, then weight assignments equally within each model. Table~\ref{tab:reasoning-usage} also reports total reasoning tokens and decision counts per dialogue. The 95\% confidence intervals use 5,000 paired cluster-bootstrap resamples, clustering by Sotopia scenario (14), ESC card (331), P4G persona (200), or CB listing (85).

To compare the same decision range, we pair dialogues by assignment, role, and repeat and retain pairs reaching at least three decisions in both conditions. Table~\ref{tab:reasoning-first-three} reports mean tokens per decision over the first three decisions, weighting eligible dialogue pairs equally.

\noindent\textbf{Cross-turn phrase reuse.} We Unicode-normalize and lowercase saved reasoning, removing function words and common interface terms. Public replies, partner messages, and prompts are excluded. Four-grams are overlapping sequences of four consecutive words in the filtered text and do not span turns. Let $G_t$ be the multiset of four-grams in turn $t$ and $H_{<t}$ the set of distinct four-grams in the agent's earlier reasoning. The reuse rate is
\begin{equation}
R_t=\frac{\sum_{g\in G_t}\mathbf{1}[g\in H_{<t}]}{|G_t|}.
\end{equation}

\begin{figure*}[!tp]
\centering
\includegraphics[width=\textwidth]{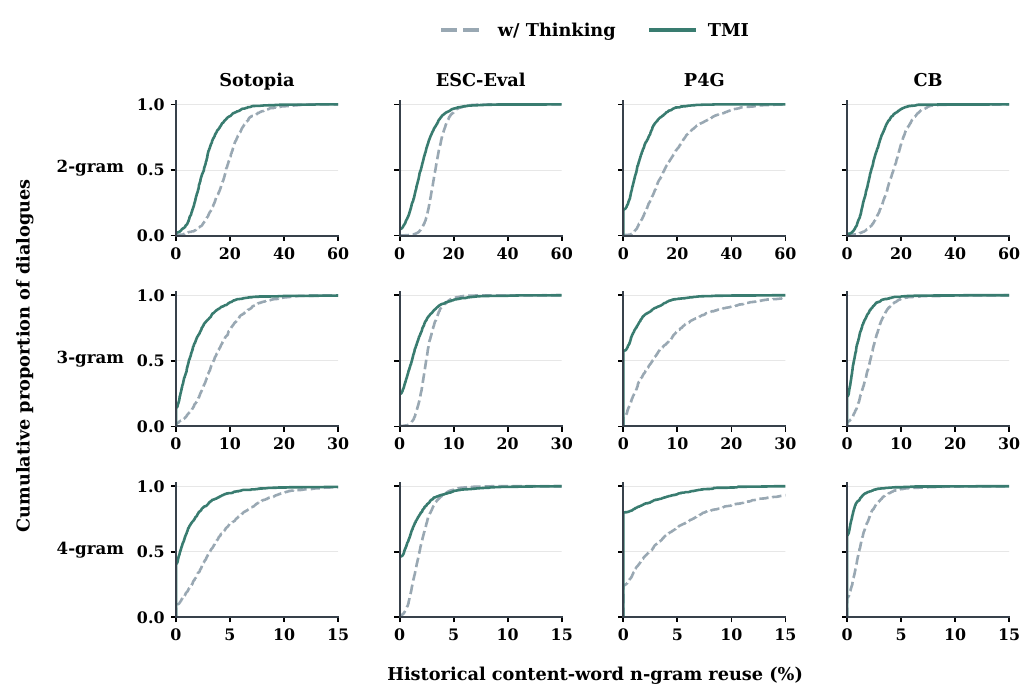}
\caption{Historical phrase reuse for $n=2,3,4$, using the same paired prefixes as Figure~\ref{fig:reasoning-length-repetition}c--f. Models receive equal weights; display limits retain full ECDF denominators.}
\label{fig:reasoning-repetition-ngram}
\end{figure*}

For each dialogue pair, we retain consecutive turns from the beginning, stopping before either condition has missing reasoning or fewer than four content words. Dialogue-level reuse averages $R_t$ over retained turns from $t=2$. To obtain mean reuse, we average repeats within assignments and weight assignments equally. ECDFs weight eligible dialogues equally within each model and give both models equal weight; observations outside the displayed range remain in their denominators. Figure~\ref{fig:reasoning-repetition-ngram} uses the same paired prefixes for $n=2,3,4$.

\Needspace{4\baselineskip}
\noindent\textbf{Length control.} At each paired turn, we retain the first $K=\min(64,\ell_{\mathrm{Thinking}},\ell_{\mathrm{TMI}})$ content words from each trace, where $\ell$ is its length in content words. We apply this rule separately to every earlier turn, matching both current and historical text lengths. Reuse rates are then recomputed with the same assignment weighting. Table~\ref{tab:reasoning-reuse} reports full-text means and differences after length matching; the latter are calculated from the truncated texts, with intervals from the bootstrap procedure above.

\subsection{Reasoning-Preservation Control}
\label{app:reasoning-preservation-control}
\begin{figure*}[!tp]
\centering
\includegraphics[width=0.95\textwidth]{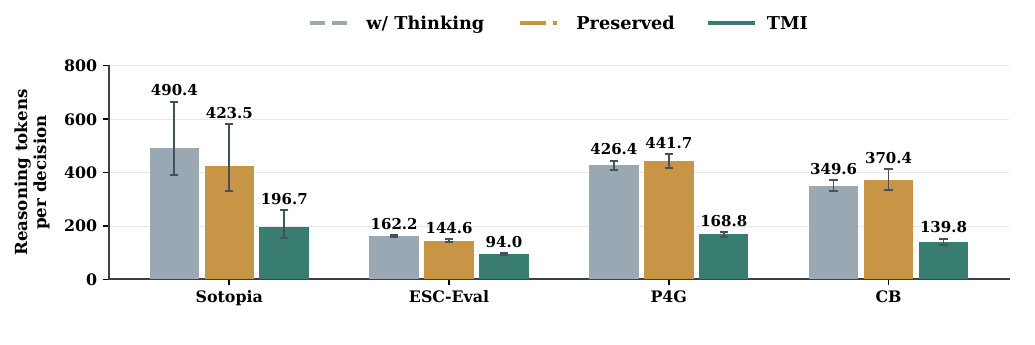}
\caption{Qwen reasoning tokens per decision under Thinking, Preserved, and TMI, with 95\% cluster-bootstrap confidence intervals.}
\label{fig:reasoning-preservation-length}
\end{figure*}

Figure~\ref{fig:reasoning-preservation-length} applies the token accounting and bootstrap procedure above to Qwen's Thinking, Preserved, and TMI conditions.

\begin{figure*}[!tp]
\centering
\includegraphics[width=0.95\textwidth]{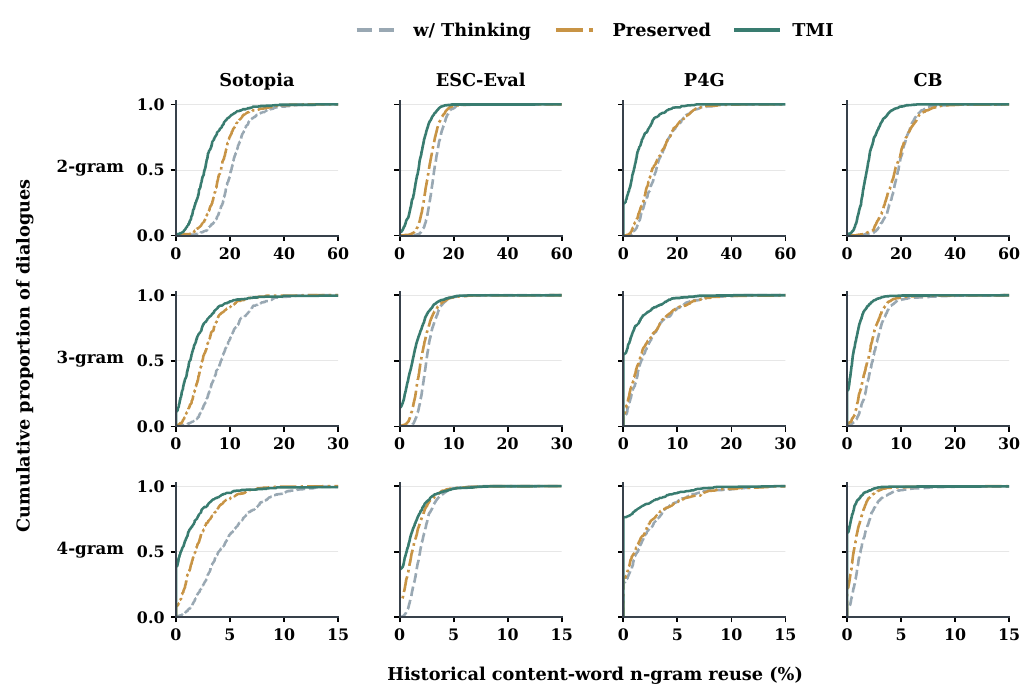}
\caption{Qwen historical n-gram reuse under Thinking, Preserved, and TMI, using common prefixes across all three conditions. Display limits retain full ECDF denominators.}
\label{fig:reasoning-preservation-ngram}
\end{figure*}

Phrase comparisons match assignments and repeats across all three conditions, retaining common prefixes with at least four content words per turn and two eligible turns. This yields 408 dialogues per condition in Sotopia, 987 in ESC, 525 in P4G, and 589 in CB. Figure~\ref{fig:reasoning-preservation-ngram} uses the same prefixes for $n=2,3,4$. ECDFs weight dialogues equally; reported means use the assignment weighting defined above.

\Needspace{12\baselineskip}
\section{Interaction Prompts}
\label{app:interaction-prompts}
Tables~\ref{tab:prompt-sotopia}--\ref{tab:prompt-cb} present the shared task prompts and condition-specific instructions. Braced fields denote task-specific inputs. Sotopia's \texttt{\{initial\_history\}} includes participant-visible context, profiles, and the private goal. P4G and CB append initial observations when available; ESC provides the seeker's latest reply separately. The two baselines use identical prompts and differ only in native thinking availability. Prompt wording follows the experimental inputs. The tables omit JSON output specifications and tool schemas.

\setlength{\abovecaptionskip}{5pt}
\begin{table*}[!p]
\begin{minipage}[t][\dimexpr\textheight-2pt\relax][s]{\textwidth}
\centering\fontsize{9}{9.4}\selectfont
\setlength{\tabcolsep}{4pt}\renewcommand{\arraystretch}{1.0}
\begin{tabularx}{\textwidth}{@{}p{2.5cm}X@{}}
\toprule[1.2pt]
\multicolumn{2}{@{}l@{}}{\textbf{Sotopia-Hard: shared task prompt}}\\
\midrule
\multicolumn{2}{@{}p{\textwidth}@{}}{
\begin{minipage}{\textwidth}
\fontsize{9}{9.4}\linespread{0.96}\selectfont
\setlength{\parindent}{0pt}\setlength{\parskip}{0pt}
Imagine you are \texttt{\{agent\}}, your task is to act/speak as \texttt{\{agent\}} would, keeping in mind \texttt{\{agent\}}'s social goal.\par
You can find \texttt{\{agent\}}'s goal (or background) in the 'Here is the context of the interaction' field.\par
Note that \texttt{\{agent\}}'s goal is only visible to you.\par
You should try your best to achieve \texttt{\{agent\}}'s goal in a way that align with their character traits.\par
Additionally, maintaining the conversation's naturalness and realism is essential (e.g., do not repeat what other people has already said before).\par
\texttt{\{initial\_history\}}.\par
Your available action types are\par
none speak non-verbal communication action leave.\par
\end{minipage}}\\
\midrule
\textbf{Condition} & \textbf{Condition-specific instruction excerpts}\\
\midrule
\textbf{Baseline} & Note: You can "leave" this conversation if 1. you have achieved your social goals, 2. this conversation makes you uncomfortable, 3. you find it uninteresting/you lose your patience, 4. or for other reasons you want to leave.\par\\
\midrule
\textbf{TMI} & For each turn, respond with exactly one \texttt{interact} tool call. Use the tool schema to specify your action; do not reply with text, JSON, or an explanation.\par
\vspace{1.5pt}
Note: You can call \texttt{interact} with \texttt{leave} if 1. you have achieved your social goals, 2. this conversation makes you uncomfortable, 3. you find it uninteresting/you lose your patience, 4. or for other reasons you want to leave.\par\\
\bottomrule[1.2pt]
\end{tabularx}
\caption{Sotopia-Hard task prompt and condition-specific instructions.}
\label{tab:prompt-sotopia}
\par\vfill
\centering\fontsize{9}{9.4}\selectfont
\setlength{\tabcolsep}{4pt}\renewcommand{\arraystretch}{1.0}
\begin{tabularx}{\textwidth}{@{}p{2.5cm}X@{}}
\toprule[1.2pt]
\multicolumn{2}{@{}l@{}}{\textbf{P4G persuader: shared task prompt}}\\
\midrule
\multicolumn{2}{@{}p{\textwidth}@{}}{
\begin{minipage}{\textwidth}
\fontsize{9}{9.4}\linespread{0.96}\selectfont
\setlength{\parindent}{0pt}\setlength{\parskip}{0pt}
In the following conversation, you will play as a Persuader in a persuasion game. Your partner has earned \$2.00. They can donate any amount from \$0.00 to \$2.00 to Save the Children; the donation comes out of their payment. Your goal is to persuade your partner to donate as much as possible.\par
\vspace{1.5pt}
Related information:\par
- Save the Children is an international non-governmental organization that promotes children's rights, provides relief, and supports children in developing countries.\par
- The donation is deducted from your partner's \$2.00 payment. The research team collects all donations and sends them to Save the Children.\par
\vspace{1.5pt}
Use this information with the conversation, in your own words. Never invent facts, numbers, ratings, or sources. If you lack a detail, do not claim it; return to what you do know or to your partner's concerns.\par
\vspace{1.5pt}
Choose one strategy for each reply. Strategy is private; only Response is sent to your partner.\par
Strategies for persuasion:\par
- logical\_appeal: Give a relevant factual reason for donating.\par
- emotion\_appeal: Appeal to concern for children without inventing details.\par
- credibility\_appeal: Address the charity's trustworthiness using information you have.\par
- foot\_in\_the\_door: Invite a small commitment before a larger request, when appropriate.\par
- self\_modeling: Describe your own willingness to help without claiming an action you have not taken.\par
- personal\_story: Share a relevant experience only if it is true; do not invent one.\par
- donation\_information: Explain the amount, deduction, or how donations reach the charity.\par
- source\_related\_inquiry: Ask what your partner knows or thinks about Save the Children.\par
- task\_related\_inquiry: Ask about your partner's questions or views on this donation task.\par
- personal\_related\_inquiry: Ask about your partner's experiences or priorities related to giving.\par
- none: Use for a greeting, routine reply, donation proposal or confirmation, or a turn where no listed strategy fits.\par
\vspace{1.5pt}
Reply in one or two short, natural sentences, under 50 words.\par
\end{minipage}}\\
\midrule
\textbf{Condition} & \textbf{Condition-specific instruction excerpts}\\
\midrule
\textbf{Baseline} & only Response is shown to your partner.\par
Do not include explanations or Markdown fences.\par\\
\midrule
\textbf{TMI} & Make exactly one interact tool call following the tool schema; only Response is shown to your partner.\par
Do not send a separate text response or explanation.\par\\
\bottomrule[1.2pt]
\end{tabularx}
\caption{P4G persuader task prompt and condition-specific instructions.}
\label{tab:prompt-p4g}
\end{minipage}
\end{table*}

\begin{table*}[!p]
\begin{minipage}[t][\dimexpr\textheight-2pt\relax][s]{\textwidth}

\centering\fontsize{9}{9.4}\selectfont
\setlength{\tabcolsep}{4pt}\renewcommand{\arraystretch}{1.0}
\begin{tabularx}{\textwidth}{@{}p{2.5cm}X@{}}
\toprule[1.2pt]
\multicolumn{2}{@{}l@{}}{\textbf{ESC-Eval supporter: shared task prompt}}\\
\midrule
\multicolumn{2}{@{}p{\textwidth}@{}}{
\begin{minipage}{\textwidth}
\fontsize{9}{9.4}\linespread{0.96}\selectfont
\setlength{\parindent}{0pt}\setlength{\parskip}{0pt}
You are a supportive and empathetic assistant for an emotional support context.\par
\vspace{1.5pt}
Strategies for emotional support:\par
- question: Ask for details to better understand the seeker's situation and emotions.\par
- restatement: Paraphrase the situation or experience described by the seeker.\par
- reflection: Reflect the seeker's feelings about their situation.\par
- self\_disclosure: Share a relevant personal experience to show empathy only when grounded in your assigned identity or background. Do not invent lived experiences.\par
- affirmation: Affirm the seeker's strengths, motivation, and capabilities; offer reassurance and encouragement.\par
- suggestions: Offer practical advice or actionable steps relevant to the seeker's situation.\par
- information: Provide factual information relevant to the seeker's situation.\par
- others: Exchange pleasantries or use support strategies outside the categories above.\par
\vspace{1.5pt}
Respond in a calm and non-judgmental manner, acknowledging and validating the seeker's emotions.\par
Ask gentle clarifying questions only when helpful, and offer practical coping suggestions when appropriate.\par
Do not diagnose, prescribe, or claim professional authority.\par
Keep responses concise and natural (2-4 sentences).\par
\end{minipage}}\\
\midrule
\textbf{Condition} & \textbf{Condition-specific instruction excerpts}\\
\midrule
\textbf{Baseline} & only Response is shown to the seeker.\par
Do not include explanations or Markdown fences.\par\\
\midrule
\textbf{TMI} & Make exactly one interact tool call following the tool schema; only Response is shown to the seeker.\par
Do not send a separate text response or explanation.\par
The tool returns the seeker's next reply, or a notice that the interaction has ended.\par\\
\bottomrule[1.2pt]
\end{tabularx}
\caption{ESC-Eval supporter task prompt and condition-specific instructions.}
\label{tab:prompt-esc}
\par\vfill

\centering\fontsize{9}{9.4}\selectfont
\setlength{\tabcolsep}{4pt}\renewcommand{\arraystretch}{1.0}
\begin{tabularx}{\textwidth}{@{}p{2.5cm}X@{}}
\toprule[1.2pt]
\multicolumn{2}{@{}l@{}}{\textbf{CB buyer: shared task prompt}}\\
\midrule
\multicolumn{2}{@{}p{\textwidth}@{}}{
\begin{minipage}{\textwidth}
\fontsize{9}{9.4}\linespread{0.96}\selectfont
\setlength{\parindent}{0pt}\setlength{\parskip}{0pt}
You are the Buyer negotiating to purchase \texttt{\{title\}}.\par
Item description:\par
\texttt{\{item\_description\}}\par
\vspace{1.5pt}
The listing price is \$\texttt{\{listing\_price\}}. Your target price is \$\texttt{\{target\_price\}}.\par
Negotiate to get as close to your target price as possible.\par
\vspace{1.5pt}
Negotiate through ordinary conversation.\par
\vspace{1.5pt}
Transaction rules:\par
- To finalize a deal, reply with [AGREED: \$\textless price\textgreater] using a positive price with at most two decimal places.\par
- The price in AGREED must use the same pricing unit as the listing price and target price: per item if the listing is per item, or for the whole bundle if the listing is for a bundle. Do not substitute a multi-item total for a per-item price.\par
- A deal completes only when both participants confirm the same price with AGREED in consecutive messages. If you disagree with a proposed price, keep negotiating.\par
- If no agreement can be reached, reply with [NO DEAL].\par
- Write transaction markers exactly as shown, with a literal dollar sign and no backslashes.\par
- Use at most one transaction marker per reply.\par
\vspace{1.5pt}
Choose one Strategy for each reply. Strategy is private; only Response is sent to the other participant.\par
Strategies for negotiation:\par
- greet: Open with a greeting or brief small talk.\par
- inquire: Ask about the item, price, condition, or terms.\par
- inform: Answer a question or provide relevant item information you know.\par
- propose: Make the first specific price offer.\par
- counter: Make a new specific price offer after an earlier offer.\par
- counter-noprice: Discuss moving the price without stating a new number.\par
- confirm: Ask for clarification or confirmation of a term.\par
- affirm: Affirm a requested confirmation.\par
- deny: Deny a requested confirmation.\par
- agree: Accept a price proposal.\par
- disagree: Reject a price proposal without a new offer.\par
- none: Use for another natural reply or ending that does not fit the listed strategies.\par
\end{minipage}}\\
\midrule
\textbf{Condition} & \textbf{Condition-specific instruction excerpts}\\
\midrule
\textbf{Baseline} & only Response is shown to the other participant.\par
Do not include explanations or Markdown fences.\par\\
\midrule
\textbf{TMI} & Make exactly one interact tool call following the tool schema; only Response is shown to the other participant. Do not send a separate text reply.\par\\
\bottomrule[1.2pt]
\end{tabularx}
\caption{CB buyer task prompt and condition-specific instructions.}
\label{tab:prompt-cb}
\end{minipage}
\end{table*}


\end{document}